\documentclass[balance,upint,subscriptcorrection,varvw,mathalfa=cal=boondoxo,spanish,french,vietnamese,russian,greek,pdf-a,colorlinks]{asmeconf}

\hypersetup{
	pdfauthor={Zongshuo Li},
	pdftitle={Efficient Cutting Tool Wear Segmentation Based on Segment Anything Model},
	pdfkeywords={Tool wear detection, Semantic segmentation, Segment Anything Model, Computer vision, Intelligent manufacturing},
	pdfsubject = {ASME MSEC2024-124455, Zongshuo Li},
	pdflicenseurl={https://ctan.org/pkg/asmeconf},
}

\usepackage{stfloats}

\begin{document}


\ConfName{Proceedings of the ASME 2024 19th International\linebreak Manufacturing Science and Engineering Conference}
\ConfAcronym{MSEC2024}
\ConfDate{June 17 – June 21, 2024} 
\ConfCity{Knoxville, Tennessee} 
\PaperNo{MSEC2024-124455}


\title{Efficient Cutting Tool Wear Segmentation Based on Segment Anything Model} 
 
%
%
%

\SetAuthors{%
	Zongshuo Li\affil{1}\CorrespondingAuthor{z.li@mti.rwth-aachen.de}, 
	Ding Huo\affil{1}, 
	Markus Meurer\affil{1}, 
	Thomas Bergs\affil{1}\affil{2}
	}

\SetAffiliation{1}{Manufacturing Technology Institute (MTI) of RWTH Aachen University, Aachen, Germany}
\SetAffiliation{2}{Fraunhofer Institute for Production Technology IPT, Aachen, Germany}


\maketitle





\begin{abstract}
Tool wear conditions impact the surface quality of the workpiece and its final geometric precision. In this research, we propose an efficient tool wear segmentation approach based on Segment Anything Model, which integrates U-Net as an automated prompt generator to streamline the processes of tool wear detection. Our evaluation covered three Point-of-Interest generation methods and further investigated the effects of variations in training dataset sizes and U-Net training intensities on resultant wear segmentation outcomes. The results consistently highlight our approach's advantage over U-Net, emphasizing its ability to achieve accurate wear segmentation even with limited training datasets. This feature underscores its potential applicability in industrial scenarios where datasets may be limited.
\end{abstract}


\begin{nomenclature}[2cm]
\entry{SAM}{Segment Anything Model}
\entry{PoI}{Point-of-Interest}
\entry{CV}{Computer Vision}
\entry{ML}{Machine Learning}
\entry{CNN}{Convolutional Neural Network}
\entry{FCN}{Fully Convolutional Network}
\entry{IoU}{Intersection over Union}
\entry{ReLU}{Rectified Linear Unit}
\entry{LLM}{Large Language Model}
\entry{CoG}{Center of Gravity}
\entry{MS}{Mask Shrink}
\entry{CoGA}{Center of Gravity Adjustment}
\entry{RCoGA}{Recursive Center of Gravity Adjustment}
\entry{BCE}{Binary Cross-Entropy}
\entry{Tr.}{Training Dataset}
\entry{ANOVA}{Analysis of Variance}
\end{nomenclature}


\section{Introduction}
Tool wear is an inevitable phenomenon in the actual machining process. It leads to alterations in the cutting zone's process variables like the forces and temperatures exerted on both the tool and workpiece. These conditions not only influence the rate of tool wear but also affect the surface quality and geometric precision of the workpiece~\cite{Wang2018}. Therefore, tool wear is one of the key determinants of both tool costs and the quality of the finished workpiece, emphasizing the necessity for monitoring during the machining process to ensure optimal outcomes~\cite{EZUGWU19991,Wang2018}.

Tool wear measurement is broadly classified into two categories: Direct and Indirect methods~\cite{JEON1988207}. Indirect methods use sensors, such as dynamometers, to consistently track tool wear through wear model~\cite{Abellan-Nebot2010,Zhou2018}. A fundamental requisite for indirect methods is the execution of tool wear experiments, paired with direct tool wear measurements, to comprehend the actual tool state~\cite{BERGS2020947}. Direct tool wear measurements predominantly use optical sensors, such as optical microscopes, with subsequent manual or automated tool wear detection. In data-intensive automated production systems, automated tool wear detection driven by classical Computer Vision (CV) and/or Machine Learning (ML) offer advantages over manual methods. Classical CV like Sobel and Canny algorithms, along with the active contour method~\cite{4767851,996,Kass1988}, have been extensively utilized for detecting tool wear~\cite{DAddona2017,peng_2020_study,BAGGA2021771}. Their primary benefit lies in the ability to obtain reasonable outcomes with only a minimal dataset~\cite{GARCIAPEREZ202385}. However, these methods are sensitive to challenging lighting conditions and varied background environments, complicating the robust feature extraction~\cite{xia_2022_segmentation}.

In contrast, considering the common challenges in metal cutting, such as fluctuating lighting conditions, coating colors, and variations in tool geometries, ML algorithms demonstrate superior performance in tool wear detection through their advanced adaptive capabilities and outperforms traditional image processing methods~\cite{DAddona2017,FRIEDRICH2023425,garcaords_2016_a}. Hou et al.~\cite{hou_2018_a} developed a self-matching algorithm to measure milling cutter wear, achieving a mean absolute error of less than 7 µm of maximum width of the flank wear land, and a maximum error under 57 µm for severe wear and breakage scenarios. Tool wear detection inherently aligns with semantic segmentation tasks, for which U-Net has increasingly become one of the favored solutions due to its effectiveness~\cite{RFB15a,9238462,SCHLEGEL2024118270,wu_2023_tool}.

In our previous research, we proposed a methodology based on U-Net for the classification of tool types and the detection of tool wear areas in microscopic images. This approach yielded promising outcomes, achieving a mean Intersection over \mbox{Union (IoU)} of 0.73 on the test dataset of hybrid multiple tools~\cite{BERGS2020947}. Subsequently, we developed a comprehensive pipeline integrating U-Net to process tool wear images captured with a digital microscope and employed a rule-based methodology to quantify wear along the tool cutting edge. This approach facilitates fully automated tool wear detection and measurement, showing a high alignment ($R^{2}$ = 0.99) with manual measurement benchmarks~\cite{HOLST2022534}. The application of U-Net for tool wear segmentation was further refined and advanced. Xia et al.~\cite{xia_2022_segmentation} introduced an enhanced SE-U-Net, supplementing network outputs with morphological processing to achieve an average segmentation accuracy of 92\%. Furthermore, Chen et al.~\cite{chen_2023_semantic} utilized a generative adversarial network to augment the dataset and employed transfer learning to boost the segmentation network's generalization capacity. Compared to training with only small samples, this approach improves the mean IoU by 8.97\%.

However, U-Net encounters limitations, particularly in the low accuracy of tool wear recognition for unknown and disturbed tool images. On one hand, the method needs a substantial dataset to ensure the model's generalization capacity, enabling it to make reliable predictions even with data disturbances. This challenge involves not only deviations in images of the same tool captured under varying conditions, but also deviations across tools with different geometries. On the other hand, the requirement for these substantial data volumes inherently leads to an extensive need for manual annotation~\cite{BERGS2020947}. Such demands, both in terms of dataset size and extensive labeling, are often impractical in industrial environments. For any specific tool, the number of images obtainable from experiments and suitable for training the network, is typically limited. This paucity of data leads to a limitation of this method for direct industrial applications.

In recent years, there has been notable progress in the development of pre-trained large language models (LLM). These models, trained on expansive datasets, exhibit remarkable zero-shot generalization abilities, which means they can handle previously unseen tasks without retraining. A common technique employed with these models is prompt engineering, which directs the model to address specific tasks. Drawing inspiration from LLMs, Meta introduced a promptable semantic segmentation model, Segment Anything Model (SAM), in CV. Pre-trained on an extensive dataset comprising 1 billion masks and 11 million images, this model demonstrates robust generalization~\cite{kirillov2023segment}.

Therefore, building on previous research, this study introduces a novel approach to tool wear segmentation, integrating U-Net with SAM. This approach takes advantage of SAM's robust generalization capabilities to post-process U-Net's outputs, ensuring high accuracy in tool wear segmentation even with limited datasets and the lower quality outputs of U-Net. We have also explored the impact of various factors on the final tool wear segmentation accuracy, including the methods for generating the Point-of-Interest (PoI), the volume of training data, and the \mbox{U-Net} training intensity. Chapter 2 describes our newly proposed approach. Chapter 3 details the prepared dataset, the model parameters, and the evaluation strategy. Chapter 4 is dedicated to a comprehensive analysis and discussion of results. The last chapter summarizes our work and gives an outlook on its future development.


\section{Segmentation of Tool Wear Based on SAM}
This chapter begins with an introduction to U-Net and SAM, followed by the description of our proposed approach.

\subsection{U-Net}
\begin{figure}[h]
\centering\includegraphics[width=0.9\linewidth]{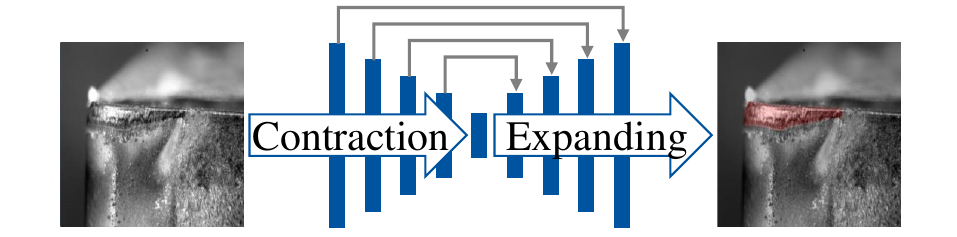}
\caption{U-Net on tool wear segmentation task}\label{fig:2_1_Unet_Example}
\end{figure}

U-Net stands out as an FCN architecture. Characterized by a substantial number of feature channels, U-Net effectively propagates contextual information to its higher resolution layers. The architecture follows a U-shaped structure, with its symmetric contraction and expanding paths. The training process of \mbox{U-Net} is end-to-end, which enables the network to produce image segmentation masks directly from the original input image~\cite{RFB15a}. U-Net's contraction path uses a classical Convolutional Neural Network (CNN) architecture. Within each stage, it incorporates two 3$\times$3 convolution units, immediately succeeded by a rectified linear unit (ReLU) activation function, and a subsequent 2$\times$2 max-pooling operation for downsampling. Conversely, in the expanding path, a 2$\times$2 up-convolution unit is employed. A special attribute of U-Net is the incorporation of skip connections, which facilitate the addition of the output from the contraction path on the same stage into the expanding path. This structural feature of combining low-level features with high-level features helps to avoid information loss~\cite{9238462}. The final layer of the network employs a 1$\times$1 convolution to map all feature channels associated with each pixel to the class corresponding to that pixel. This elegant architecture ensures its applicability extends beyond its initial biomedical segmentation domain and can be easily applied to other tasks like tool wear semantic segmentation. Figure~\ref{fig:2_1_Unet_Example} shows an example of U-Net on a tool wear segmentation task. 


\subsection{SAM}

\begin{figure}[t]
\centering\includegraphics[width=0.9\linewidth]{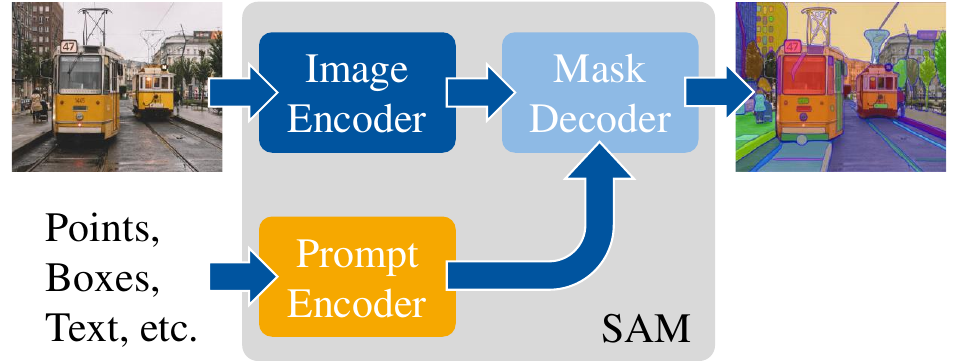}
\caption{overview of SAM's structure}\label{fig:2_2_SAM_Structure}
\end{figure}

SAM, serving as a foundation model for image segmentation, can efficiently produce accurate image segmentation masks with prompts that indicate segmentation targets in the image. The prompts can be in various formats, including target or background PoI, boxes, masks, or even free-form text~\cite{kirillov2023segment}. SAM consists of three parts: an image encoder, a prompt encoder, and a mask decoder. Figure~\ref{fig:2_2_SAM_Structure} shows an overview of SAM's structure.


The image encoder uses a vision transformer proposed by Dosovitskiy et al.~\cite{dosovitskiy2021image} and is pre-trained with a mask autoencoder~\cite{he2021masked}. Through attention blocks, the image encoder can process high-resolution inputs and produce a downscaled image embedding with dimensions of ${64^2\times256}$. This image embedding is a numerical representation of the image that captures their semantic meaning and visual features.

The prompt encoder convolves dense prompts (masks) to match the dimensionality of the image embedding and adds them element-wise up. For sparse prompts, including points and boxes, the prompt encoder maps them into a 256-dimensional vectorial embedding by positional encodings and representations indicating the prompt type. The resulting prompt embedding serve as a vectorized representation of the prompts.

The mask decoder begins with a two-layer modified transformer decoder with cross-attention mechanisms to update both the image and prompt embeddings. Subsequently, the updated image embedding is upscaled through two transposed convolutional layers, while the updated prompt embedding undergoes alignment via a three-layer multilayer perceptron (MLP) to match the upscaled image's channel dimension. The final output mask is the spatially pointwise product of the upscaled image embedding and the MLP output~\cite{kirillov2023segment}. This design allows the mask decoder to produce specific segmentations corresponding to varying prompts with the identical image embedding from the image encoder.

The prompt engineering enables SAM to be adapted for a broad range of applications beyond its initial scope with appropriate prompts~\cite{zhang2024segment,ji2023segment}. For example, Osco et al.~\cite{OSCO2023103540} demonstrated its capability in remote sensing by evaluating SAM's performance across multiscale datasets with diverse input prompts, including bounding boxes, individual points, and text.


\begin{figure*}[b]
\centering\includegraphics[width=0.9\linewidth]{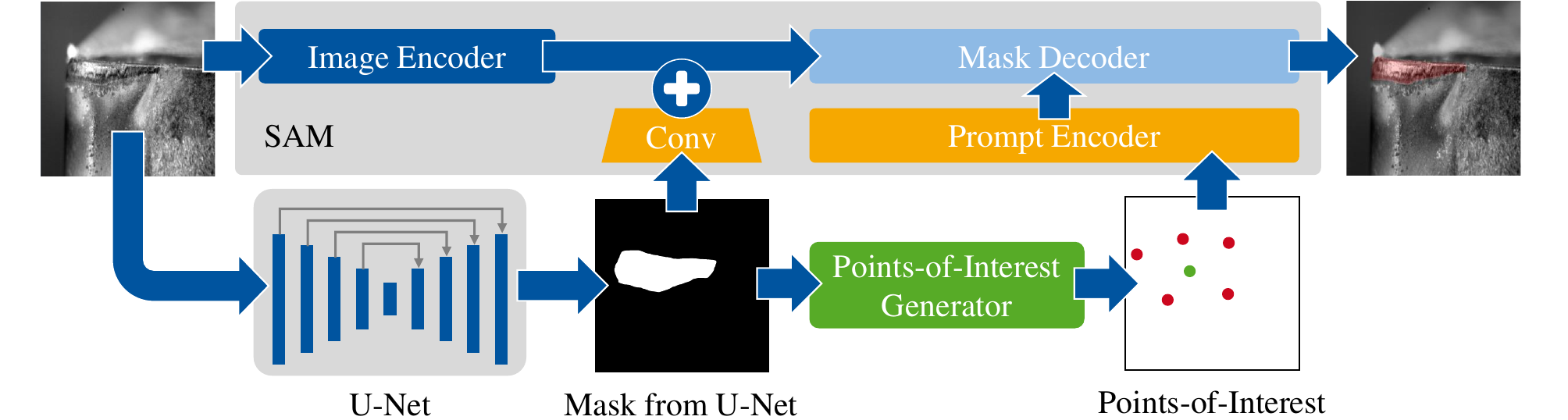}
\caption{the concept of the proposed approach}\label{fig:2_2_Our_Approach}
\end{figure*}


\subsection{Our Approach}

In our previous research, U-Net has demonstrated effective performance in specific tasks. With a training dataset comprising similar data, U-Net can achieve precise tool wear segmentation results for either individual tools or multiple tools within a composite test set~\cite{BERGS2020947}. This observation aligns with findings from other studies as well~\cite{9238462,SCHLEGEL2024118270,wu_2023_tool,xia_2022_segmentation,chen_2023_semantic}. However, a notable decline in U-Net's performance is observed when applied to tools dissimilar from those in the training set. This indicates the necessity of additional, scene-specific training datasets for each unique scenario encountered~\cite{BERGS2020947}. Therefore, the direct deployment of U-Net in an industrial context faces challenges due to limited training data and its weak generalization capability for unfamiliar data. Despite these constraints, U-Net is still capable of providing preliminary, albeit semi-accurate, tool wear segmentation. On the other hand, although SAM has the robust zero-shot generalization capability, its direct application to the tool wear segmentation task yields suboptimal performance. This shortfall is primarily due to SAM's inability to accurately segment tool wear without appropriate prompt guidance. Therefore, the use of appropriate prompts offers potential to improve SAM's performance, e.g., when combined with U-Net. Consequently, in this approach, we employ U-Net as an automated prompt generator. Figure~\ref{fig:2_2_Our_Approach} illustrates the concept of our proposed approach. Initially, U-Net provides an initial image segmentation mask. This mask undergoes binarization as the first step. Following this, employing a PoI generator, this binarized mask is used to produce a set of PoI, including target points (representing tool wear) and background points (indicating non-wear). The resulting PoI, in combination with the initial binarized mask, is processed by SAM's prompt encoder. Finally, SAM's mask decoder will process the prompts to output the precise tool wear segmentation mask.

\subsection{Point of Interest Generator}
The goal of the PoI generator is to generate positive points in the center of the foreground region of the image segmentation mask, thereby prompting the segmentation target. In this regard PoI generator also generate negative points in the background region of the mask to enhance the prompt. We employed three different methods to identify the location of the positive point. These include the Mask Shrink (MS) method~\cite{Said_2021}, the Center of Gravity Adjustment (CoGA) method, and the Recursive Center of Gravity Adjustment (RCoGA) method.

\subsubsection*{MS}
\begin{figure}[htbp]
\centering\includegraphics[width=0.9\linewidth]{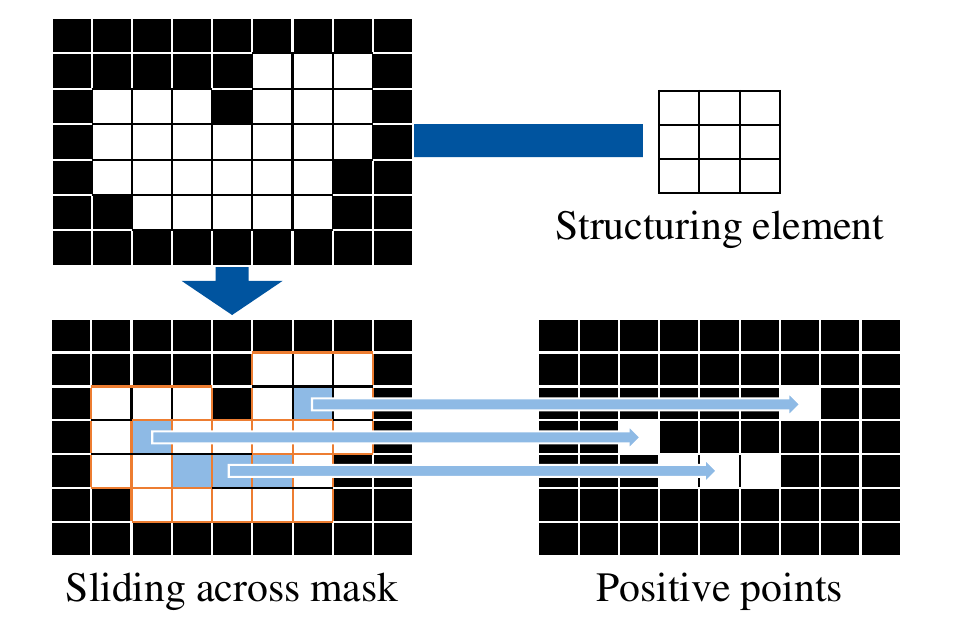}
\caption{procedural flow of MS}\label{fig:2_3_MS}
\end{figure}

The MS method is derived from morphological image processing, aimed at enhancing image quality by shrinking the foreground region through successive erosion operations~\cite{Said_2021}. MS can also deduce the central point or line of a mask following multiple iterations of shrinking. Figure~\ref{fig:2_3_MS} shows the procedural flow of MS. To initiate, an appropriate structuring element is chosen. This element is then systematically slid across the input mask. If the mask segment beneath the structuring element does not align with it, that segment is designated as the background. Conversely, pixels located at the central position of the structuring element are assigned as the foreground, with the remaining as the background. Successive repetitions of this operation yield the image’s central point or line.


\subsubsection*{CoGA}
\begin{figure}[htbp]
\centering\includegraphics[width=0.9\linewidth]{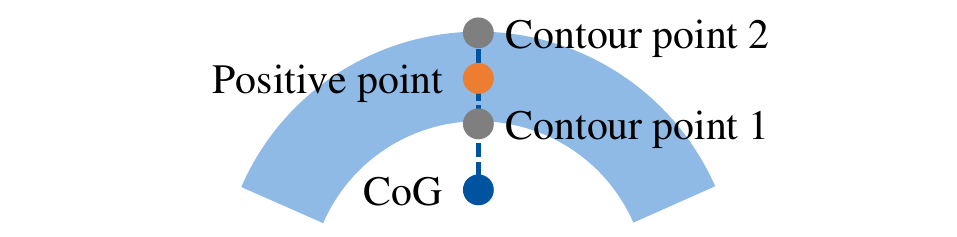}
\caption{procedural flow of CoGA}\label{fig:2_3_CoGA}
\end{figure}

We propose a simple method for determining the single center of gravity (CoG), aiming for a fast identification of the positive point within the foreground region. Figure~\ref{fig:2_3_CoGA} shows the procedural flow of CoGA. Initially, the mask's foreground CoG is calculated. If this CoG locates within the mask region, it is directly returned as the positive point. Conversely, if it locates outside, the nearest point on the foreground contour (contour point 1) to this CoG is identified. Drawing a line between this CoG and contour point 1 intersects the foreground region at another contour point 2, producing a line segment contour point 1 to 2. The midpoint of this segment is then returned as the final positive point.


\subsubsection*{RCoGA}
\begin{figure}[htbp]
\centering\includegraphics[width=0.9\linewidth]{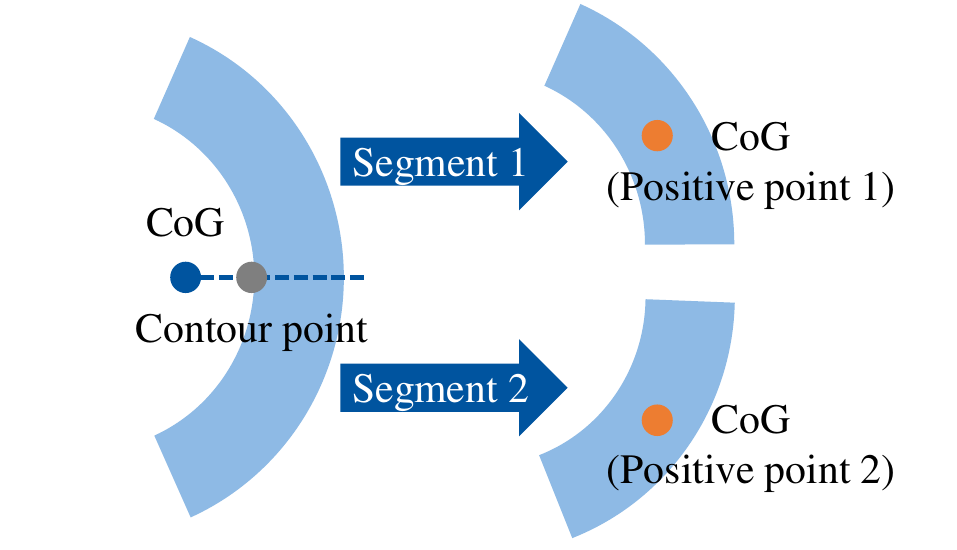}
\caption{procedural flow of RCoGA}\label{fig:2_3_RCoGA}
\end{figure}

Based on CoGA, we propose RCoGA as a method to generate multiple positive points. This approach aims to counteract potential prompt deficiencies in single positive point derived from CoGA.  Figure~\ref{fig:2_3_RCoGA} shows the procedural flow of RCoGA. Initially, the CoG of the mask's foreground is computed. Same as CoGA, if this CoG locates within the mask region, it is directly returned as the positive point. Otherwise, the nearest point on the foreground contour (contour point) to this CoG is identified. Drawing a line through this CoG and this contour point cuts the foreground into several smaller segments. This CoG calculation and cutting process is recursively executed on each segment until every segment returns a positive point. The depth of this recursive process can be controlled by capping the number of iterations.


\subsubsection*{Generation of Negative Points}
\begin{figure}[htbp]
\centering\includegraphics[width=0.9\linewidth]{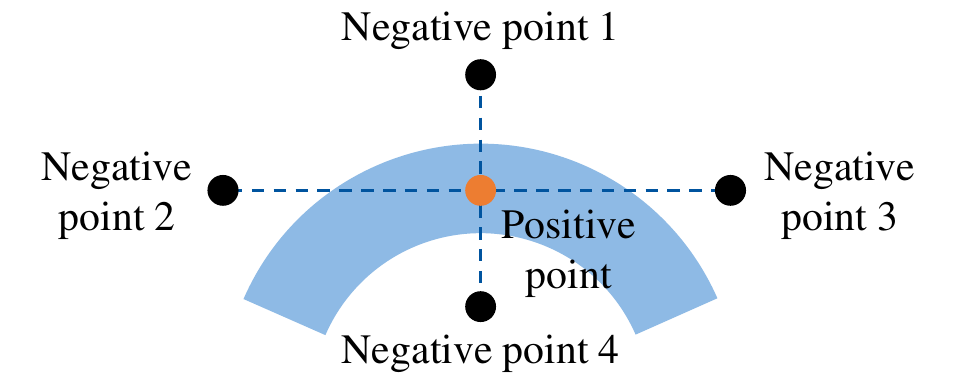}
\caption{Generation of Negative points}\label{fig:2_3_NP}
\end{figure}

Building upon the identified positive points, we find four contour points on the foreground contour, horizontally and vertically for each positive point. Subsequently, four extrapolated points are derived by extending a specified pixel distance along the direction from the positive point to the contour points. If these extrapolated points locate within the foreground mask, they are dropped. Otherwise, they are returned as negative points. Figure~\ref{fig:2_3_NP} shows the negative points generation process.


\section{Materials and Methods}
This chapter details the prepared dataset, the model parameters, and the evaluation strategy.

\subsection{Dataset}
\begin{figure}[htbp]
\centering
\begin{subfigure}[t]{0.25\linewidth} 
\centering
\includegraphics[width=0.9\linewidth]{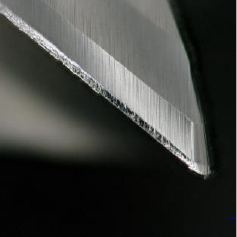}
\subcaption{End Mill 1\label{fig:3_1_Sample_Dataset_1}}
\end{subfigure}%
\begin{subfigure}[t]{0.25\linewidth}
\centering
\includegraphics[width=0.9\linewidth]{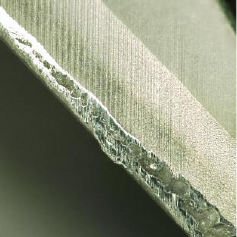}
\subcaption{End Mill 2\label{fig:3_1_Sample_Dataset_2}}
\end{subfigure}%
\begin{subfigure}[t]{0.25\linewidth}
\centering
\includegraphics[width=0.9\linewidth]{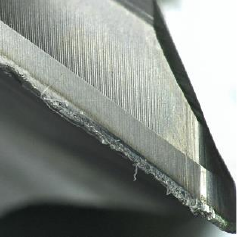}
\subcaption{End Mill 3\label{fig:3_1_Sample_Dataset_3}}
\end{subfigure}\\%
\begin{subfigure}[t]{0.25\linewidth} 
\centering
\includegraphics[width=0.9\linewidth]{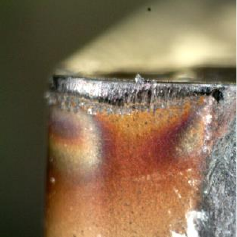}
\subcaption{Insert 1\label{fig:3_1_Sample_Dataset_4}}
\end{subfigure}%
\begin{subfigure}[t]{0.25\linewidth}
\centering
\includegraphics[width=0.9\linewidth]{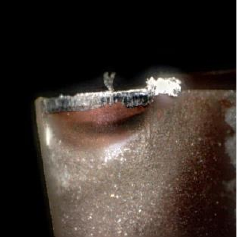}
\subcaption{Insert 2\label{fig:3_1_Sample_Dataset_5}}
\end{subfigure}%
\begin{subfigure}[t]{0.25\linewidth}
\centering
\includegraphics[width=0.9\linewidth]{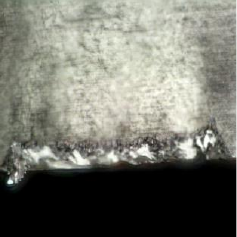}
\subcaption{Insert 3\label{fig:3_1_Sample_Dataset_6}}
\end{subfigure}%
\caption{Sample of different tools in the dataset}\label{fig:3_1_Sample_Dataset}
\end{figure}

In this study, the image dataset was sourced from micrographs of the flank face of cutting tools, captured utilizing a Keyence VHX-6000 series microscope in prior experimental trials. Six different tools, including different end mills and inserts, are included in this dataset. They vary in coating, substrate, material being machined and process. This provides a comprehensive view of various tool types and their wear patterns. The focus of this study was specifically on flank wear. These images are stored in an RGB three-channel format, with a consistent resolution of 1024 x 1024 pixels. Within the dataset corresponding to each tool, 20\% of the images are stochastically selected as the test dataset, with selection criteria predicated upon wear area distribution. This selection ensures homogeneity in wear area distribution between both training and test datasets. Model performance evaluation will be conducted on this designated test dataset to endure the comparability between different models. The residual 80\% constitutes the training dataset. A detailed enumeration of tool types and the corresponding number of images is presented in Table~\ref{tab:dataset}. Additionally, Figure~\ref{fig:3_1_Sample_Dataset} exhibits a selection of representative images.


\begin{table}[htbp]
\caption[Table]{Specifications of training and test dataset}\label{tab:dataset}
\centering{%
\begin{tabular}{llr}
\toprule
Tool types & Training dataset & Test dataset\\
\midrule
End Mill 1	& 177   &  37 \\
End Mill 2  & 207   &  45 \\
End Mill 3	& 90    &  19 \\
Insert 1 	& 192   &  37 \\
Insert 2 	& 192   &  40 \\
Insert 3 	& 54 	&  12 \\
\bottomrule
\end{tabular}
}
\end{table}


Data augmentation plays an important role in enhancing the network's robustness, especially in scenarios with limited training samples~\cite{DBLP:journals/corr/abs-1806-03852}. Through data augmentation, there is not only an expansion in the volume of images with diversified distributions during model training, but also an enhancement in model generalization and a consequential reduction in overfitting~\cite{DBLP:journals/corr/abs-1712-04621}. Given the limited quantity of available images in this study, we have incorporated specific data augmentation operations. Table~\ref{tab:augmentation} delineates those operations.


\begin{table}[b]
\caption[Table]{Specifications of augmentation operations}\label{tab:augmentation}
\centering{%
\begin{tabular}{lr}
\toprule
Augmentation operations & Settings \\
\midrule
Horizontal flip 	& probability of 50\%	\\
Vertical flip		& probability of 50\%	\\
Rotate 				& -20 to 20 degrees	\\
Translate			& -10\% to 10\%		\\
\bottomrule
\end{tabular}
}
\end{table}

\begin{figure}[htbp]
\centering
\begin{subfigure}[t]{0.3\linewidth} 
\centering
\includegraphics[width=0.9\linewidth]{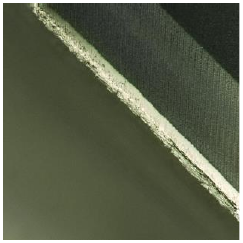}
\subcaption{Original Image\label{fig:3_1_Label_Example_1}}
\end{subfigure}%
\begin{subfigure}[t]{0.3\linewidth}
\centering
\includegraphics[width=0.9\linewidth]{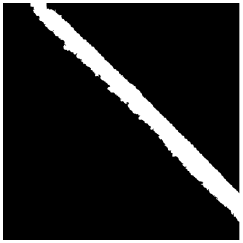}
\subcaption{Binarized Label\label{fig:3_1_Label_Example_2}}
\end{subfigure}%
\begin{subfigure}[t]{0.3\linewidth}
\centering
\includegraphics[width=0.9\linewidth]{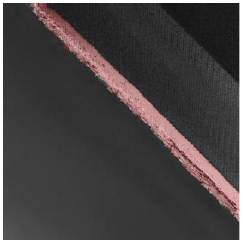}
\subcaption{Overlaid Result\label{fig:3_1_Label_Example_3}}
\end{subfigure}%
\caption{Example of pixel-level labeling}\label{fig:3_1_Label_Example}
\end{figure}

Pixel-level labeling is necessary for training semantic segmentation models. Therefore, all images are manually labeled with worn regions. The resultant label is a binarized image, wherein each pixel is categorized as either 'worn' or 'non-worn'. Figure~\ref{fig:3_1_Label_Example} presents an example: the original image of an end mill, the corresponding binarized labels, and an overlaid result of them.


\subsection{Metrics}
The IoU metric, also known as Jaccard index~\cite{f6902328-1c27-32dc-a3cd-6b0b3a4279b1}, stands as a prevalent performance metric in image segmentation tasks. IoU computes the ratio of the intersecting area between the model's segmentation and the ground truth to the cumulative area of both within the image, with the Eq.~\eqref{eqn:iou}, where $N$ is the number of pixels, $g_i$ is the ground truth, and $y_i$ is the model output.

\begin{equation}\label{eqn:iou}
\mathrm{IoU} = \frac{2\sum_{i=1}^{N}y_ig_i }{\sum_{i=1}^{N}(y_i+g_i)} 
\end{equation}

\begin{figure}[htbp]
\centering\includegraphics[width=0.9\linewidth]{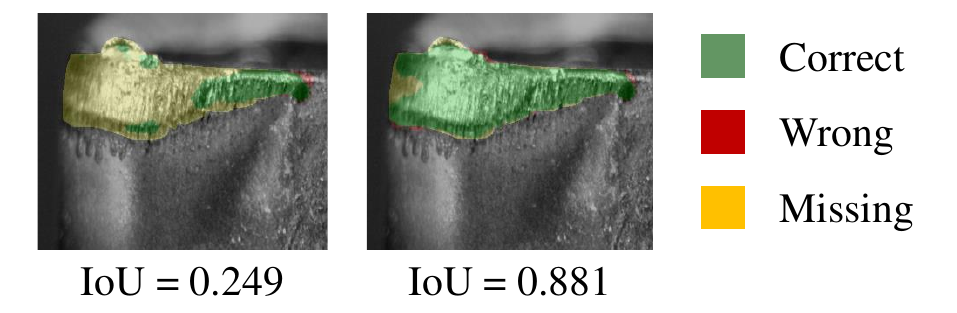}
\caption{examples of IoU values}\label{fig:3_2_IoU_Sample}
\end{figure}

IoU's value spans from 0 to 1: a value of 0 signifies no overlapping area, while a value of 1 indicates perfect match between the model's segmentation and the ground truth. Figure~\ref{fig:3_2_IoU_Sample} offers examples of IoU values for some sample segmentation results: Red denotes area where the model's segmentation is exclusive of the ground truth, green signifies correct segmentation area and yellow represents area the model failed to segment.


\subsection{U-Net Parameter}
The foundational architecture of original U-Net~\cite{RFB15a} is preserved in our implementation, maintaining the same convolutional units positioned along both the contracting and expanding paths, as well as the sampling unit. Given that our input consists of an image dimensioned at 1024$\times$1024$\times$3, the resolution for each convolutional unit within the network has been amplified by a factor of 4. Table~\ref{tab:Unet_structure} shows the implemented U-Net’s structure.


\begin{table}[htbp]
\caption[Table]{The implemented U-Net’s structure}\label{tab:Unet_structure}
\centering{%
\begin{tabular}{llr}
\toprule
Layer & Convolution & Sampling \\
\midrule
Input 	& 1024$\times$1024$\times$3		& --\\
Down1	& 2$\times$ReLU(Conv3-64)		& Max-Pooling-2\\
Down2	& 2$\times$ReLU(Conv3-128)		& Max-Pooling-2\\
Down3	& 2$\times$ReLU(Conv3-256)		& Max-Pooling-2\\
Down4	& 2$\times$ReLU(Conv3-512)		& Max-Pooling-2\\
Center	& 2$\times$ReLU(Conv3-1024)		& --\\
Up1		& 2$\times$ReLU(Conv3-512)		& Up Sampling\\
Up2		& 2$\times$ReLU(Conv3-256)		& Up Sampling\\
Up3		& 2$\times$ReLU(Conv3-128)		& Up Sampling\\
Up4		& 2$\times$ReLU(Conv3-64)		& Up Sampling\\
Output	& Softmax(Conv1-2)				& --\\
\bottomrule
\end{tabular}
}
\end{table}

Furthermore, we have modified the loss function to incorporate both binary cross-entropy (BCE) loss and IoU loss. The pixel-level BCE loss is typically employed in training networks for standard image segmentation tasks where the misclassification of fewer pixels only has minimal impact. The entire loss function, employed throughout the network’s training phase, is shown in Eq.~\eqref{eqn:loss_entire}.

\begin{equation}\label{eqn:loss_entire}
\begin{aligned}
loss= 	& l_{\mathrm{BCE}} + l_{\mathrm{IoU}} \\
	= 	& -\sum_{i=1}^{N}(g_i\log{y_i} +(1-g_i)\log{(1-y_i)}) \\
		& + 1-\frac{2\sum_{i=1}^{N}y_ig_i }{\sum_{i=1}^{N}(y_i+g_i)} 
\end{aligned}
\end{equation}

\subsection{SAM Parameter}
We retained the original configuration of the SAM without modifications and training. We selected the ViT-L SAM~\cite{kirillov2023segment} in our implementation. Given that the U-Net’s output size is 1024$\times$1024, a 2$\times$2 max-pooling operation was employed to reduce the dimension to 256$\times$256, which matches the SAM’s prompt encoder input. From the three potential masks produced by the mask decoder, the mask with the highest probability was selected as the model's final output.


\begin{figure*}[b]
\centering\includegraphics[width=0.9\linewidth]{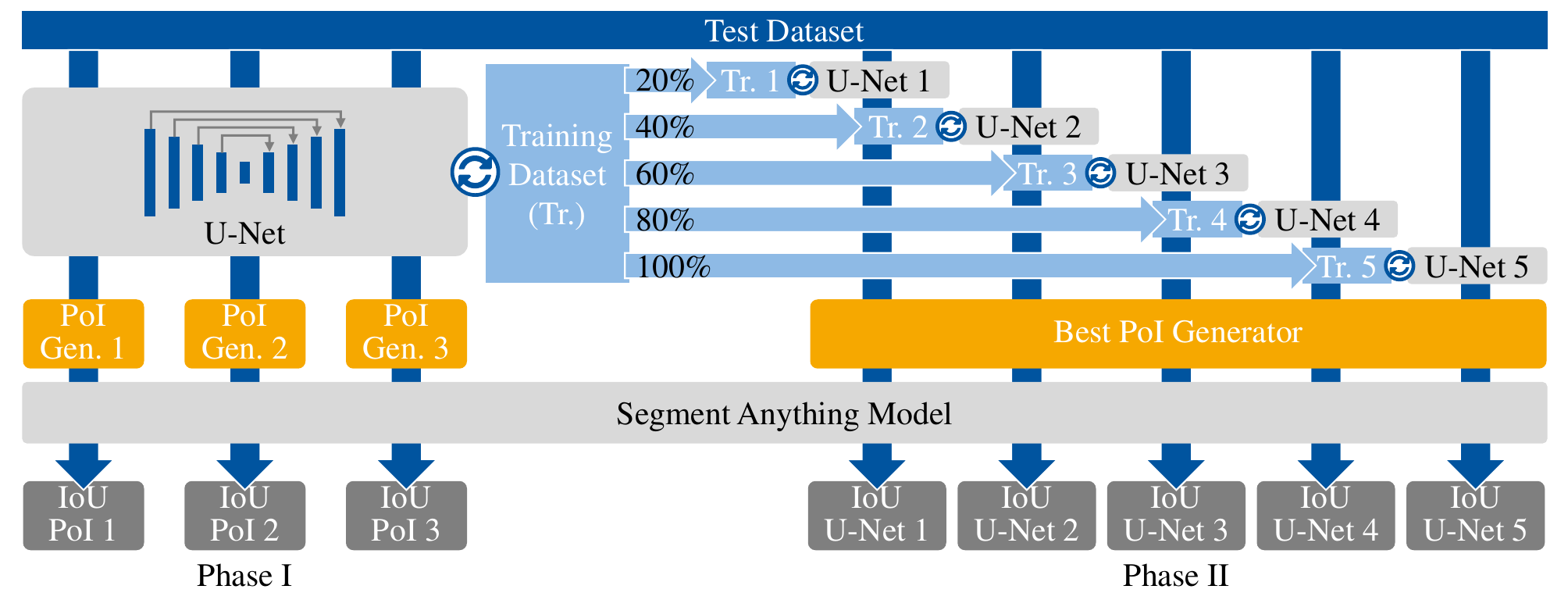}
\caption{The entire evaluation strategy}\label{fig:3_6_Evaluation_Strategy}
\end{figure*}
 

\subsection{Evaluation Strategy}

The generation of expansive datasets and their subsequent manual annotation are often impractical in industrial environments, limiting the direct industrial application of U-Net. Given this challenge, this study aims to ascertain the feasibility of the proposed approach in achieving tool wear segmentation, even with limited dataset volumes.

In the initial phase, several representative PoI generation methods were compared. The U-Net model is initially trained utilizing the entire training dataset. Subsequently, its segmentation outputs on the test dataset will serve as foundational data for PoI generation, with methods delineated in Chapter 2. These generated PoI, together with the U-Net’s image segmentation outputs, serve as input prompts for the SAM model. A comparative analysis is then conducted to evaluate SAM’s image segmentation performance relative to different PoI generation methods.

During the second phase, subsets of the original training dataset, ranging from 20\% to 100\%, are stochastically selected to constitute new training datasets of varied sizes. Several U-Nets are trained on these varied-size training datasets. PoI generation based on the outputs of U-Nets on the test dataset are then be executed, with the optimal method identified during the first phase. These points and segmentation outputs of U-Net serve as input prompts for SAM. A comparative assessment of SAM's image segmentation performance relative to U-Net’s is undertaken to evaluate the proposed approach’s feasibility in tool wear segmentation with reduced dataset sizes. Figure~\ref{fig:3_6_Evaluation_Strategy} shows the entire evaluation strategy.


\section{Results and Discussion}
This chapter initiates with a comparative analysis of the performance metrics of three distinct PoI generation methods. Subsequently, the most efficacious PoI generation method will be employed to examine the impact of training data volume on the performance of our proposed approach. Ultimately, we will assess the impact of varying U-Net training intensities on the proposed approach's performance, particularly with limited training data.

\subsection{Impact of the Points of Interest Generator (Phase I)}
\begin{figure}[htbp]
\centering\includegraphics[width=0.9\linewidth]{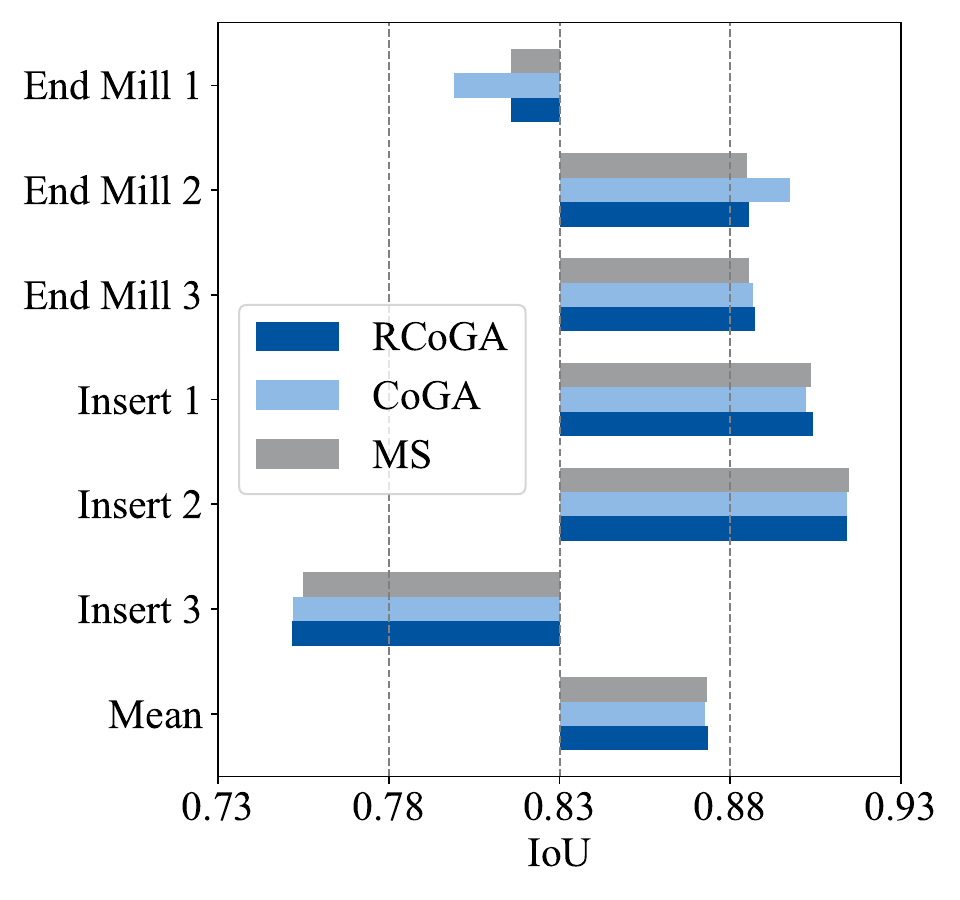}
\caption{IOU of the proposed approach on single tools using different PoI generators}\label{fig:4_1_Barh_PoI}
\end{figure}

\begin{figure}[htbp]
\centering\includegraphics[width=0.9\linewidth]{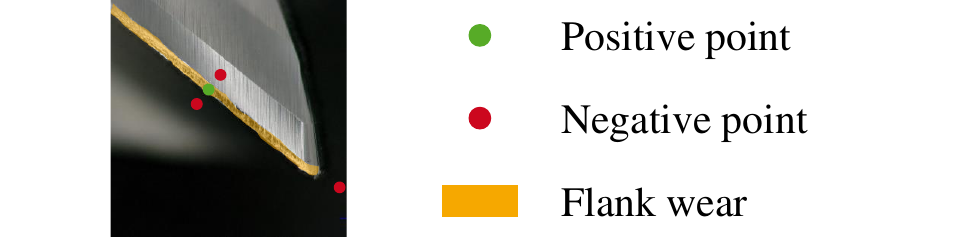}
\caption{Example of PoI generation for End Mill 1 with CoGA method}\label{fig:4_1_Endmill1_Sample}
\end{figure}

Figure~\ref{fig:4_1_Barh_PoI} shows the performance of three different PoI generation methods: MS, CoGA and RCoGA, evaluated on the test dataset. The U-Net underwent training for 100 epochs utilizing the entire training dataset and achieved convergence. As shown in Fig.~\ref{fig:4_1_Barh_PoI}, all three PoI generation methods exhibit comparable performance across the entire test dataset, each yielding an IoU of about 0.87. However, a variance in performance is observed across the test datasets for different tools. Specifically, End Mill 1 and Insert 3 exhibit suboptimal outcomes relative to the other four tools. This disparity is attributed to End Mill 1’s wear area, characterized by its small and elongated nature, leading to a single positive point that proves less effective in prompting distal wear areas. Figure~\ref{fig:4_1_Endmill1_Sample} shows an example of PoI generation for End Mill 1 with CoGA method. The results of other methods are similar. The performance on Insert 3 is compromised by the presence of unclean cutting edges like built-up edges and residual chips, impacting the model's ability to accurately segment the real wear region. Notably, in the case of End Mill 2, CoGA has an IoU advantage of 0.01 over its counterparts, while it incurs a deficit of 0.01 in End Mill 1. An Analysis of Variance (ANOVA) conducted for the PoI generation methods yielded a p-value of 0.999, indicating an absence of significant performance differences among the three methods.

\begin{figure}[htbp]
\centering\includegraphics[width=0.9\linewidth]{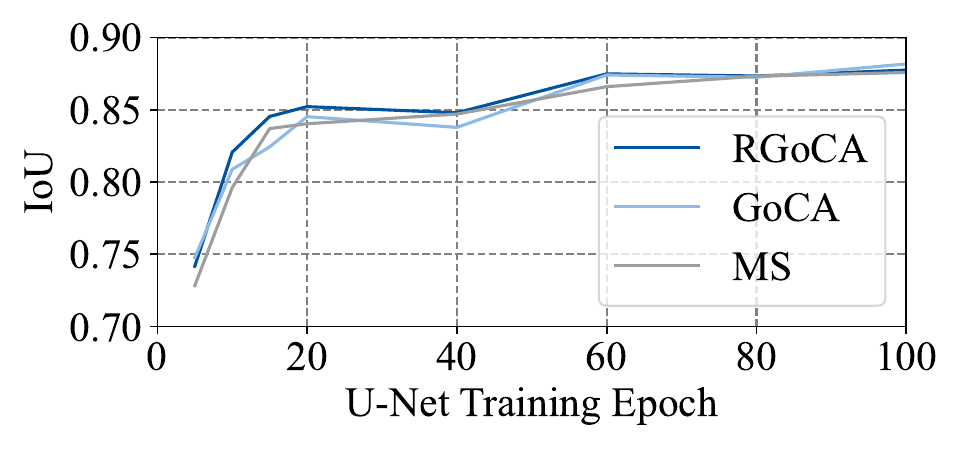}
\caption{IOU of the proposed Approach on different U-Net training epoch using different PoI generators}\label{fig:4_1_Total_PoI}
\end{figure}

Figure~\ref{fig:4_1_Total_PoI} shows the performance of the proposed approach, employing three PoI generation methods, across varied U-Net training intensities. At training intensities of both 80 and 100 epochs, all three methods exhibit near-identical performance. However, for epochs below 60, RCoGA consistently matches or outperforms both CoGA and MS, registering an approximate 0.03 advantage in comparison to the least effective method. This suggests that in case of rougher tool wear segmentations, RCoGA's strategy of generating multiple positive points by segmenting finer tool wear regions helps to provide better prompt quality. Consequently, RCoGA has been selected for PoI generation subsequent analyses.


\subsection{Impact of the Training Dataset Size (Phase II)}

\begin{figure}[htbp]
\centering\includegraphics[width=0.9\linewidth]{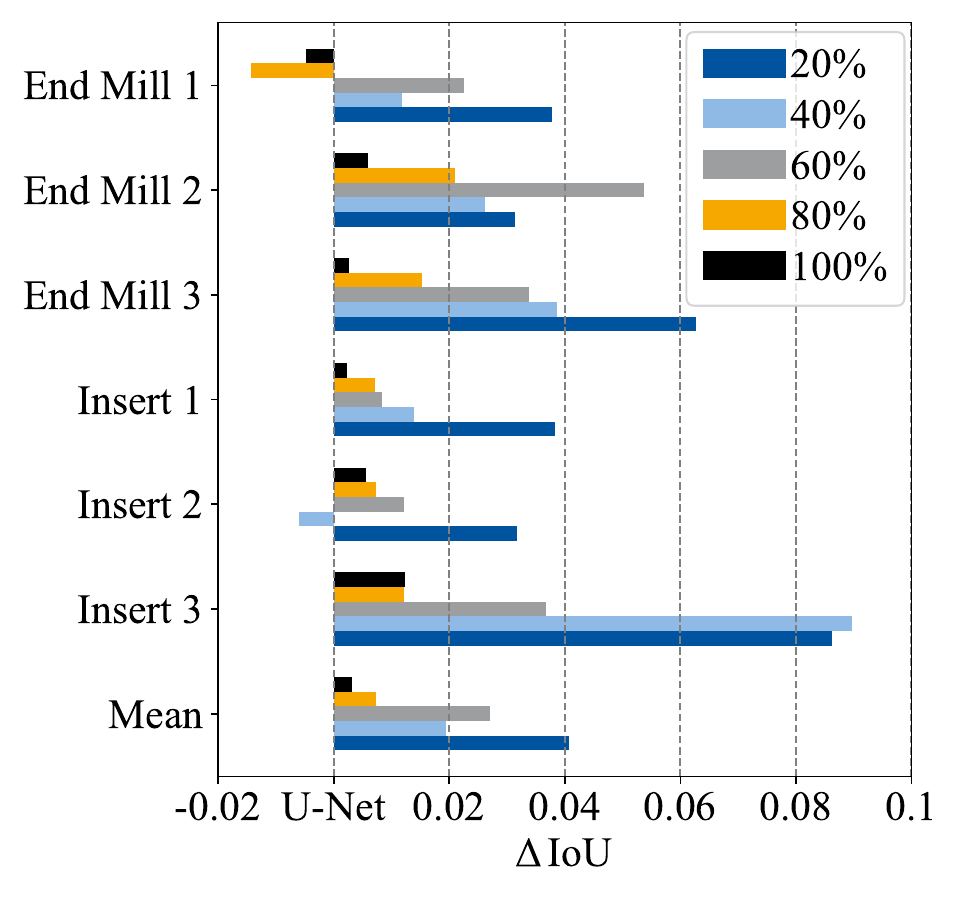}
\caption{IOU difference between the proposed Approach and U-Net on different training data volume}\label{fig:4_2_Training_Data_minus_unet}
\end{figure}

Figure~\ref{fig:4_2_Training_Data_minus_unet} shows the performance differences between U-Net and our proposed approach, setting U-Net's results as the referential baseline. U-Net undergoes training on datasets of varying sizes, with the PoI generated by RCoGA. The generated prompt from different U-Nets is fed into SAM to deliver the final tool wear segmentation outcomes. 


\begin{figure}[htbp]
\centering
\begin{subfigure}[t]{0.3\linewidth} 
\centering
\includegraphics[width=0.9\linewidth]{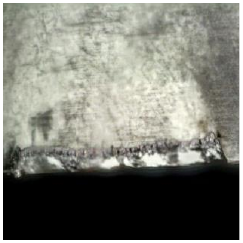}
\subcaption{Original Image\label{fig:4_2_Example_Orignal}}
\end{subfigure}%
\begin{subfigure}[t]{0.3\linewidth}
\centering
\includegraphics[width=0.9\linewidth]{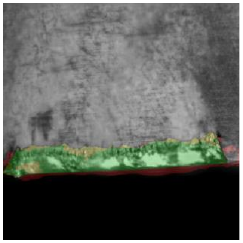}
\subcaption{U-Net\label{fig:4_2_Example_UNet}}
\end{subfigure}%
\begin{subfigure}[t]{0.3\linewidth}
\centering
\includegraphics[width=0.9\linewidth]{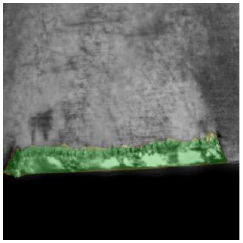}
\subcaption{Prop. Approach\label{fig:4_2_Example_SAM}}
\end{subfigure}%
\caption{Example of performance advantage}\label{fig:4_2_Example}
\end{figure}

As shown in Fig.~\ref{fig:4_2_Training_Data_minus_unet}, our proposed approach consistently demonstrates better performance relative to U-Net. The performance difference is especially prominent when using a small training dataset (20\%), with a difference ranging between 0.03 and 0.09. However, as the dataset's magnitude escalates, this advantage diminishes incrementally. For a 60\% training dataset, the advantage narrows to a range of 0.02 to 0.05, and with a full 100\% training dataset, it reduces further to less than 0.01. This performance advantage is particularly evident in the case of Insert 3, with a representative example illustrated in Fig.~\ref{fig:4_2_Example}. In this example, the IoU of the U-Net's segmentation result is only 0.65, while our approach achieves an IoU of 0.88. This outperformance can be attributed to U-Net's susceptibility to image blurring, while our approach exhibits enhanced robustness to blurring. A subsequent ANOVA conducted for training dataset size yielded a p-value of 0.001, indicating a statistically significant impact of the volume of training dataset on the observed performance discrepancy.


\begin{figure}[htbp]
\centering\includegraphics[width=0.9\linewidth]{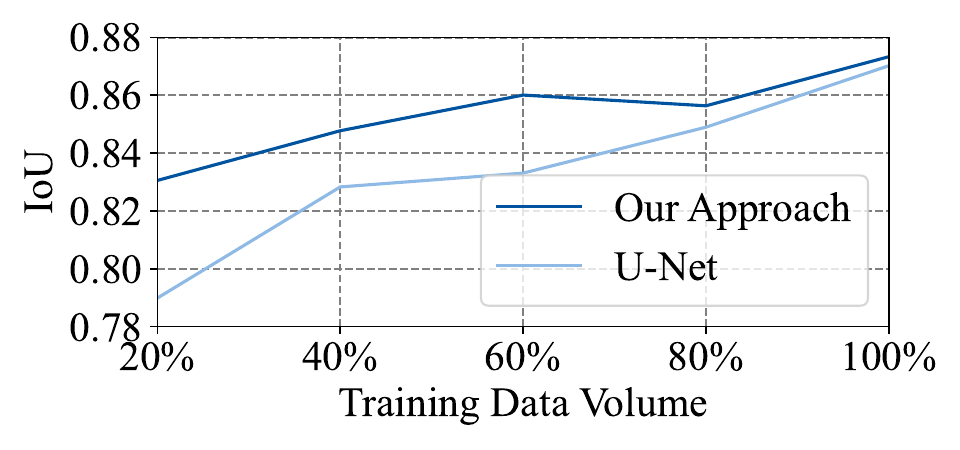}
\caption{IOU of the proposed Approach and U-Net on different training data volume}\label{fig:4_2_Training_Data_Mean}
\end{figure}

Figure~\ref{fig:4_2_Training_Data_Mean} shows the performance of both U-Net and our proposed approach on the entire test dataset across the different training dataset sizes. The results clearly indicate a consistent IoU advantage for our approach, particularly pronounced at 20\% training dataset, where it reaches an advantage of 0.04. This advantage persists up to the 60\% training dataset. From 80\% onwards, the advantage narrows to a margin of less than 0.01. Notably, our approach registers an IoU of 0.83 at a 20\% training dataset. This metric attains on U-Net only when it trained with a more substantial 60\% dataset. Furthermore, our approach outperformances U-Net with a 40\% training dataset compared to U-Net's 80\% dataset. On a 40\% training dataset, our approach already achieves a higher IoU even when U-Net is trained with a larger 80\% dataset. These observations emphasize that our proposed approach efficiently leverages smaller training datasets to match or even exceed the performance levels that U-Net achieves only with considerably larger datasets. Crucially, our approach consistently retains its performance advantage over U-Net, irrespective of training dataset size.


\subsection{Impact of the U-Net Training Intensity}

\begin{figure}[hbp]
\centering\includegraphics[width=0.9\linewidth]{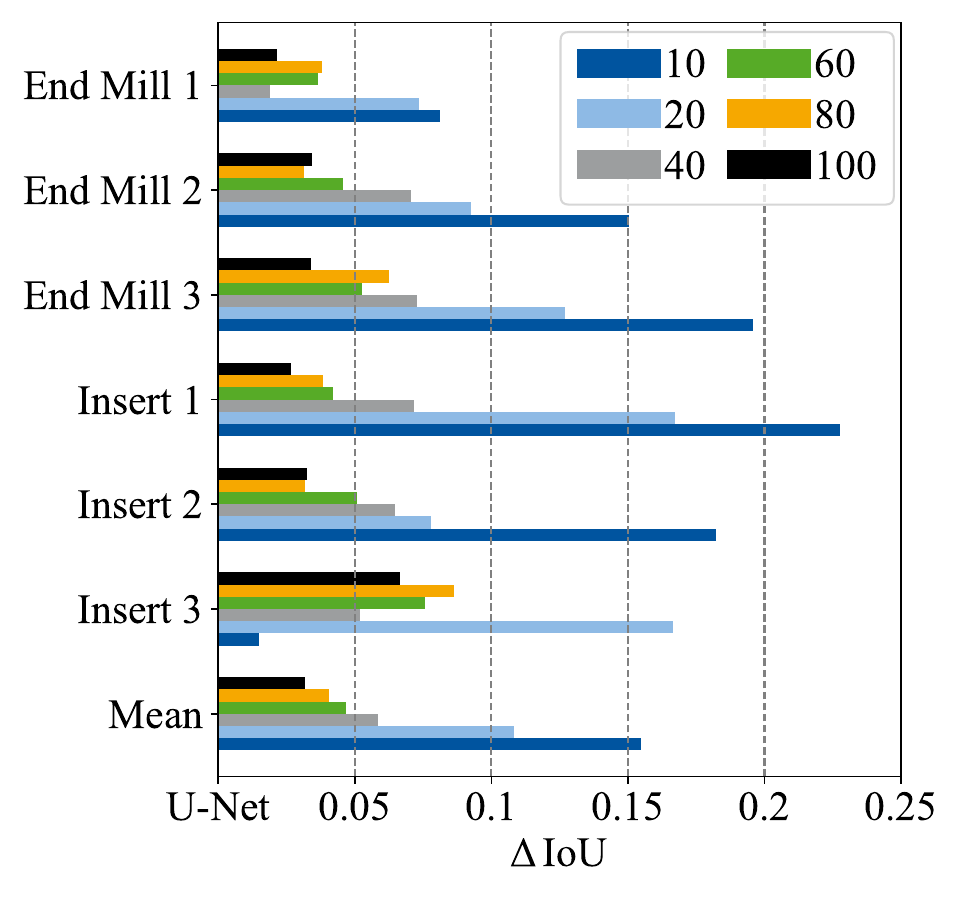}
\caption{IOU difference between the proposed Approach and U-Net on different U-Net training epoch}\label{fig:4_3_Training_Steps_minus_unet}
\end{figure}

Our approach has demonstrated pronounced advantages with limited training datasets. Based on this, we delve deeper into examining the impact of various U-Net training intensities on our approach when training with a 20\% training dataset. This simulates the performance of our approach at initial segmentations with lower quality. Figure~\ref{fig:4_3_Training_Steps_minus_unet} contrasts the performance between U-Net and our approach, with U-Net's performance serving as the reference baseline. Different colors represent various U-Net training epochs. The output of U-Net on the test dataset, across these epochs, undergoes PoI generation with RCoGA and is subsequently processed by SAM to produce the final tool wear segmentation results. As depicted, our approach exhibits substantial dominance over U-Net, accentuated at U-Net’s lower training epochs. With merely 10 epochs of U-Net training, our approach achieves an average advantage of 0.16, peaking to an advantage exceeding 0.23. At 20 epochs, this average reduces to 0.11, and further reduces to 0.06 at 40 epochs. This advantage incrementally narrows with more advanced epochs, settling between 0.03 and 0.07 at the 100-epoch. An ANOVA conducted on U-Net's training epochs returned a p-value of $3.7\times 10^{-6}$, indicating the significant impact of U-Net's training intensities on performance discrepancies. A notable exception arises with the tool Insert 3 at the 10-epoch mark, where the advantage is only 0.02. This can be attributed to U-Net's worse tool wear segmentation performance at this epoch, registering a low IoU of 0.37. Such low-quality segmentations introduce flawed prompts, detrimentally impacting our approach's efficacy.


\begin{figure}[htbp]
\centering\includegraphics[width=0.9\linewidth]{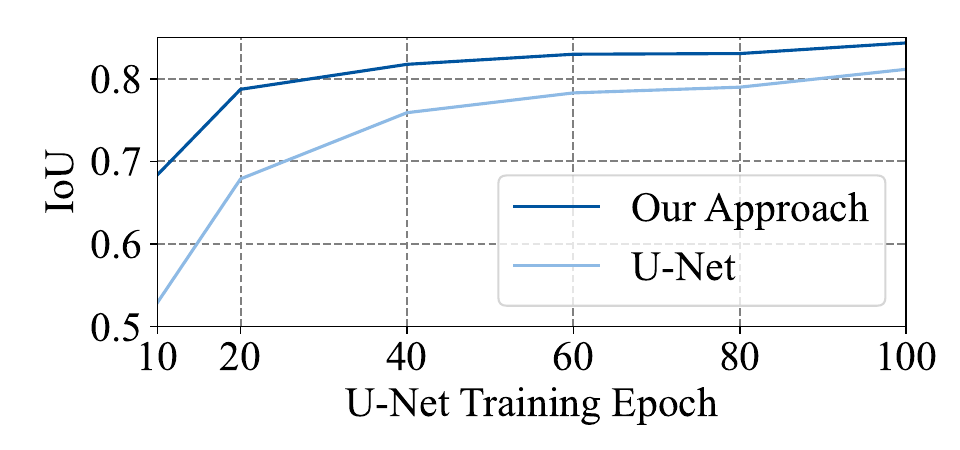}
\caption{IOU of the proposed Approach and U-Net on different U-Net training epoch}\label{fig:4_3_Training_Steps_Mean_unet}
\end{figure}

Figure~\ref{fig:4_3_Training_Steps_Mean_unet} shows the performance of both U-Net and our proposed approach on the entire test dataset across the different U-Net's training epochs. Our approach consistently outperforms U-Net. Similar conclusions can be drawn that our approach consistently outperforms U-Net, although this advantage reduces as U-Net undergoes extended training. Notably, after 20 epochs, our approach can already achieve an IoU nearing 0.8, with only a marginal improvement of 0.04 observed at the 100 epochs. Such performance underscores our approach's capacity to substantially refine U-Net’s initial rough tool wear segmentation outputs, even with limited training datasets.


\section{Conclusion and Outlook}
In this study, we propose an efficient tool wear segmentation approach based on SAM and U-Net. Our approach employs U-Net as an automated prompt generator for SAM, subsequently leveraging SAM's capabilities for accurate tool wear segmentation. An evaluation of three PoI generation methods revealed RCoGA as the optimal choice. Further analyses, benchmarked against U-Net’s performance, evaluated the impacts of variations in training dataset sizes and U-Net training intensities on the final segmentation outcomes. The results show that our approach consistently outperformed U-Net, particularly when working with smaller training datasets and at initial stages of U-Net training. These results underscore our approach's capability to deliver accurate tool wear segmentations even with limited training datasets, indicating its potential for swift integration into industrial practices with limited datasets.

It is imperative to emphasize that our work represents a preliminary attempt to employ SAM for tool wear segmentation. The use of U-Net for preliminary wear segmentation demonstrates one of many potential approaches, with alternatives including manual prompts or object detection models that may better meet the zero-shot requirements of industrial scenarios. Moreover, to optimize SAM's zero-shot performance specifically for tool wear segmentation, SAM itself needs to be retrained within this domain. Therefore, future research will encompass assessing model performance with expansive datasets and exploring the potential for retraining and transferring SAM to enhance its accuracy and robustness specifically for the tool wear segmentation domain.


\section*{Acknowledgments}
The authors appreciate the funding of this work by the Deutsche Forschungsgemeinschaft (DFG) – German Research Foundation for the project 509813741 “Entwicklung und Erforschung eines lernfähigen Systems zur Werkzeugverschleißüberwachung auf Basis künstlicher Intelligenz”.


\nocite{*}

\bibliographystyle{asmeconf}  
\bibliography{asmeconf-sample}

\begin{thebibliography}{10}
\newcommand{\enquote}[1]{``#1''}
\providecommand{\url}[1]{\texttt{#1}}
\providecommand{\urlprefix}{URL }
\expandafter\ifx\csname urlstyle\endcsname\relax
  \providecommand{\doi}[1]{DOI \discretionary{}{}{}#1}\else
  \providecommand{\doi}{DOI \discretionary{}{}{}\begingroup
  \urlstyle{rm}\Url}\fi
\providecommand{\eprint}[2][]{\urlprefix\url{#1#2}}

\bibitem{Wang2018}
Wang, Bing and Liu, Zhanqiang.
\newblock \enquote{Influences of tool structure, tool material and tool wear on
  machined surface integrity during turning and milling of titanium and nickel
  alloys: a review.}
\newblock \textit{The International Journal of Advanced Manufacturing
  Technology} Vol.~98 No.~1 (2018): pp. 1925--1975.
\newblock \doi{10.1007/s00170-018-2314-1}.

\bibitem{EZUGWU19991}
Ezugwu, E.O., Wang, Z.M. and Machado, A.R.
\newblock \enquote{The machinability of nickel-based alloys: a review.}
\newblock \textit{Journal of Materials Processing Technology} Vol.~86 No.~1
  (1999): pp. 1--16.
\newblock \doi{10.1016/S0924-0136(98)00314-8}.

\bibitem{JEON1988207}
Jeon, J.U. and Kim, S.W.
\newblock \enquote{Optical flank wear monitoring of cutting tools by image
  processing.}
\newblock \textit{Wear} Vol. 127 No.~2 (1988): pp. 207--217.
\newblock \doi{10.1016/0043-1648(88)90131-7}.

\bibitem{Abellan-Nebot2010}
Abellan-Nebot, Jose~Vicente and Subirón, Fernando~Romero.
\newblock \enquote{A review of machining monitoring systems based on artificial
  intelligence process models.}
\newblock \textit{The International Journal of Advanced Manufacturing
  Technology} Vol.~47 No.~1 (2010): pp. 237--257.
\newblock \doi{10.1007/s00170-009-2191-8}.

\bibitem{Zhou2018}
Zhou, Yuqing and Xue, Wei.
\newblock \enquote{Review of tool condition monitoring methods in milling
  processes.}
\newblock \textit{The International Journal of Advanced Manufacturing
  Technology} Vol.~96 No.~5 (2018): pp. 2509--2523.
\newblock \doi{10.1007/s00170-018-1768-5}.

\bibitem{BERGS2020947}
Bergs, Thomas, Holst, Carsten, Gupta, Pranjul and Augspurger, Thorsten.
\newblock \enquote{Digital image processing with deep learning for automated
  cutting tool wear detection.}
\newblock \textit{Procedia Manufacturing} Vol.~48 (2020): pp. 947--958.
\newblock \doi{10.1016/j.promfg.2020.05.134}.
\newblock 48th SME North American Manufacturing Research Conference, NAMRC 48.

\bibitem{4767851}
Canny, John.
\newblock \enquote{A Computational Approach to Edge Detection.}
\newblock \textit{IEEE Transactions on Pattern Analysis and Machine
  Intelligence} Vol. PAMI-8 No.~6 (1986): pp. 679--698.
\newblock \doi{10.1109/TPAMI.1986.4767851}.

\bibitem{996}
Kanopoulos, N., Vasanthavada, N. and Baker, R.L.
\newblock \enquote{Design of an image edge detection filter using the Sobel
  operator.}
\newblock \textit{IEEE Journal of Solid-State Circuits} Vol.~23 No.~2 (1988):
  pp. 358--367.
\newblock \doi{10.1109/4.996}.

\bibitem{Kass1988}
Kass, Michael and Terzopoulos, Demetri.
\newblock \enquote{Snakes: Active contour models.}
\newblock \textit{International Journal of Computer Vision} Vol.~1 No.~4
  (1988): pp. 321--331.
\newblock \doi{10.1007/BF00133570}.

\bibitem{DAddona2017}
D’Addona, D.M., Ullah, A. M. M.~Sharif and Matarazzo, D.
\newblock \enquote{Tool-wear prediction and pattern-recognition using
  artificial neural network and DNA-based computing.}
\newblock \textit{Journal of Intelligent Manufacturing} Vol.~28 No.~6 (2017):
  pp. 1285--1301.
\newblock \doi{10.1007/s10845-015-1155-0}.

\bibitem{peng_2020_study}
Peng, Ruitao, Pang, Haolin, Jiang, Haojian and Hu, Yunbo.
\newblock \enquote{Study of Tool Wear Monitoring Using Machine Vision.}
\newblock \textit{Automatic Control and Computer Sciences} Vol.~54 (2020): pp.
  259--270.
\newblock \doi{10.3103/s0146411620030062}.

\bibitem{BAGGA2021771}
Bagga, P.J., Makhesana, M.A., Patel, Kavan and Patel, K.M.
\newblock \enquote{Tool wear monitoring in turning using image processing
  techniques.}
\newblock \textit{Materials Today: Proceedings} Vol.~44 (2021): pp. 771--775.
\newblock \doi{10.1016/j.matpr.2020.10.680}.
\newblock International Conference on Materials, Processing and
  Characterization.

\bibitem{GARCIAPEREZ202385}
García-Pérez, Alberto, Ziegenbein, Amina, Schmidt, Eric, Shamsafar, Faranak,
  Fernández-Valdivielso, Asier, Llorente-Rodríguez, Raúl and Weigold,
  Matthias.
\newblock \enquote{CNN-based in situ tool wear detection: A study on model
  training and data augmentation in turning inserts.}
\newblock \textit{Journal of Manufacturing Systems} Vol.~68 (2023): pp. 85--98.
\newblock \doi{10.1016/j.jmsy.2023.03.005}.

\bibitem{xia_2022_segmentation}
Xia, Linzhi, Shi, Yizhu, Lin, Hongjie, Zheng, Houyuan, Cao, Xincheng, Chen,
  Binqiang, Zhou, Yuqing and Sun, Weifang.
\newblock \enquote{Segmentation and quantitative evaluation for tool wear
  condition via an improved SE-U-Net.}
\newblock \textit{The International Journal of Advanced Manufacturing
  Technology}  (2022)\doi{10.1007/s00170-022-09338-2}.

\bibitem{FRIEDRICH2023425}
Friedrich, Markus, Gerber, Theresa, Dumler, Jonas and Döpper, Frank.
\newblock \enquote{A system for automated tool wear monitoring and
  classification using computer vision.}
\newblock \textit{Procedia CIRP} Vol. 118 (2023): pp. 425--430.
\newblock \doi{10.1016/j.procir.2023.06.073}.
\newblock 16th CIRP Conference on Intelligent Computation in Manufacturing
  Engineering.

\bibitem{garcaords_2016_a}
García-Ordás, María~Teresa, Alegre, Enrique, González-Castro, Víctor and
  Alaiz-Rodríguez, Rocío.
\newblock \enquote{A computer vision approach to analyze and classify tool wear
  level in milling processes using shape descriptors and machine learning
  techniques.}
\newblock \textit{The International Journal of Advanced Manufacturing
  Technology} Vol.~90 (2016): pp. 1947--1961.
\newblock \doi{10.1007/s00170-016-9541-0}.

\bibitem{hou_2018_a}
Hou, Qiulin, Sun, Jie and Huang, Panling.
\newblock \enquote{A novel algorithm for tool wear online inspection based on
  machine vision.}
\newblock \textit{The International Journal of Advanced Manufacturing
  Technology} Vol. 101 (2018): pp. 2415--2423.
\newblock \doi{10.1007/s00170-018-3080-9}.

\bibitem{RFB15a}
Ronneberger, Olaf, Fischer, Philipp and Brox, Thomas.
\newblock \enquote{U-Net: Convolutional Networks for Biomedical Image
  Segmentation.} (2015).
\newblock \eprint[https://arxiv.org/abs/]{1505.04597}.

\bibitem{9238462}
Miao, Huihui, Zhao, Zhibin, Sun, Chuang, Li, Bing and Yan, Ruqiang.
\newblock \enquote{A U-Net-Based Approach for Tool Wear Area Detection and
  Identification.}
\newblock \textit{IEEE Transactions on Instrumentation and Measurement} Vol.~70
  (2021): pp. 1--10.
\newblock \doi{10.1109/TIM.2020.3033457}.

\bibitem{SCHLEGEL2024118270}
Schlegel, Clemens, Molitor, Dirk~Alexander, Kubik, Christian, Martin,
  Daniel~Michael and Groche, Peter.
\newblock \enquote{Tool wear segmentation in blanking processes with fully
  convolutional networks based digital image processing.}
\newblock \textit{Journal of Materials Processing Technology} Vol. 324 (2024):
  p. 118270.
\newblock \doi{10.1016/j.jmatprotec.2023.118270}.

\bibitem{wu_2023_tool}
Wu, Cheng and Wang, Shenlong.
\newblock \enquote{Tool wear assessment and life prediction model based on
  image processing and deep learning.}
\newblock \textit{The International Journal of Advanced Manufacturing
  Technology} Vol. 126 (2023): pp. 1303--1315.
\newblock \doi{10.1007/s00170-023-11189-4}.

\bibitem{HOLST2022534}
Holst, Carsten, Yavuz, Taha~Berk, Gupta, Pranjul, Ganser, Philipp and Bergs,
  Thomas.
\newblock \enquote{Deep learning and rule-based image processing pipeline for
  automated metal cutting tool wear detection and measurement.}
\newblock \textit{IFAC-PapersOnLine} Vol.~55 No.~2 (2022): pp. 534--539.
\newblock \doi{10.1016/j.ifacol.2022.04.249}.
\newblock 14th IFAC Workshop on Intelligent Manufacturing Systems IMS 2022.

\bibitem{chen_2023_semantic}
Chen, Chang, Lin, Chen, Meng, Zhen, Ni, Jing, Sun, Jiteng and Li, Zuji.
\newblock \enquote{Semantic segmentation of end mill wear area based on
  transfer learning with small dataset.}
\newblock \textit{The International Journal of Advanced Manufacturing
  Technology} Vol. 127 (2023): pp. 3599--3609.
\newblock \doi{10.1007/s00170-023-11725-2}.

\bibitem{kirillov2023segment}
Kirillov, Alexander, Mintun, Eric, Ravi, Nikhila, Mao, Hanzi, Rolland, Chloe,
  Gustafson, Laura, Xiao, Tete, Whitehead, Spencer, Berg, Alexander~C., Lo,
  Wan-Yen, Dollár, Piotr and Girshick, Ross.
\newblock \enquote{Segment Anything.} (2023).
\newblock \eprint[https://arxiv.org/abs/]{2304.02643}.

\bibitem{dosovitskiy2021image}
Dosovitskiy, Alexey, Beyer, Lucas, Kolesnikov, Alexander, Weissenborn, Dirk,
  Zhai, Xiaohua, Unterthiner, Thomas, Dehghani, Mostafa, Minderer, Matthias,
  Heigold, Georg, Gelly, Sylvain, Uszkoreit, Jakob and Houlsby, Neil.
\newblock \enquote{An Image is Worth 16x16 Words: Transformers for Image
  Recognition at Scale.} (2021).
\newblock \eprint[https://arxiv.org/abs/]{2010.11929}.

\bibitem{he2021masked}
He, Kaiming, Chen, Xinlei, Xie, Saining, Li, Yanghao, Dollár, Piotr and
  Girshick, Ross.
\newblock \enquote{Masked Autoencoders Are Scalable Vision Learners.} (2021).
\newblock \eprint[https://arxiv.org/abs/]{2111.06377}.

\bibitem{zhang2024segment}
Zhang, Yichi, Shen, Zhenrong and Jiao, Rushi.
\newblock \enquote{Segment Anything Model for Medical Image Segmentation:
  Current Applications and Future Directions.} (2024).
\newblock \eprint[https://arxiv.org/abs/]{2401.03495}.

\bibitem{ji2023segment}
Ji, Wei, Li, Jingjing, Bi, Qi, Liu, Tingwei, Li, Wenbo and Cheng, Li.
\newblock \enquote{Segment Anything Is Not Always Perfect: An Investigation of
  SAM on Different Real-world Applications.} (2023).
\newblock \eprint[https://arxiv.org/abs/]{2304.05750}.

\bibitem{OSCO2023103540}
Osco, Lucas~Prado, Wu, Qiusheng, {de Lemos}, Eduardo~Lopes, Gonçalves,
  Wesley~Nunes, Ramos, Ana Paula~Marques, Li, Jonathan and Marcato, José.
\newblock \enquote{The Segment Anything Model (SAM) for remote sensing
  applications: From zero to one shot.}
\newblock \textit{International Journal of Applied Earth Observation and
  Geoinformation} Vol. 124 (2023): p. 103540.
\newblock \doi{10.1016/j.jag.2023.103540}.

\bibitem{Said_2021}
Said, K A~M and Jambek, A~B.
\newblock \enquote{Analysis of Image Processing Using Morphological Erosion and
  Dilation.}
\newblock \textit{Journal of Physics: Conference Series} Vol. 2071 No.~1
  (2021): p. 012033.
\newblock \doi{10.1088/1742-6596/2071/1/012033}.

\bibitem{DBLP:journals/corr/abs-1806-03852}
Hernández-García, Alex and König, Peter.
\newblock \enquote{Data augmentation instead of explicit regularization.}
  (2020).
\newblock \eprint[https://arxiv.org/abs/]{1806.03852}.

\bibitem{DBLP:journals/corr/abs-1712-04621}
Perez, Luis and Wang, Jason.
\newblock \enquote{The Effectiveness of Data Augmentation in Image
  Classification using Deep Learning.} (2017).
\newblock \eprint[https://arxiv.org/abs/]{1712.04621}.

\bibitem{f6902328-1c27-32dc-a3cd-6b0b3a4279b1}
Jaccard, Paul.
\newblock \enquote{The Distribution of the Flora in the Alpine Zone.}
\newblock \textit{The New Phytologist} Vol.~11 No.~2 (1912): pp. 37--50.
\newblock Accessed 2023-10-17,
  \urlprefix\url{http://www.jstor.org/stable/2427226}.

\end{thebibliography}

\end{document}